\documentclass{article}

\usepackage{graphicx}
\usepackage{enumitem}
\setlist[enumerate]{leftmargin=*}
\usepackage{listings}
\usepackage{tabularx}
\usepackage{url}

\usepackage[preprint]{custom}

\usepackage[numbers]{natbib}
\usepackage[utf8]{inputenc} 
\usepackage[T1]{fontenc}    
\usepackage{booktabs}       
\usepackage{amsfonts}       
\usepackage{nicefrac}       
\usepackage{xcolor}         
\usepackage{subcaption}

\usepackage{hyperref}
\hypersetup{breaklinks=true}
\usepackage[protrusion=true,expansion=false]{microtype}

\title{Ensuring Reproducibility in Generative AI Systems for General Use Cases: A Framework for Regression Testing and Open Datasets}

\author{%
  Masumi Morishige \quad
  Ryo Koshihara \\[2pt]
  Galirage Inc.\\
  \texttt{info@galirage.com}
}

\date{\today}

\begin{document}

\maketitle


\begin{abstract}
Reproducibility and reliability remain pressing challenges for generative AI systems whose behavior can drift with each model update or prompt revision. We introduce \textbf{GPR-bench}, a lightweight, extensible benchmark that operationalizes regression testing for general purpose use cases. GPR-bench couples an open, bilingual (English and Japanese) dataset covering eight task categories (e.g., text generation, code generation, and information retrieval) and 10 scenarios in each task categories (80 total test cases for each language) with an automated evaluation pipeline that employs ``LLM-as-a-Judge'' scoring of \emph{correctness} and \emph{conciseness}. Experiments across three recent model versions—\texttt{gpt-4o-mini}, \texttt{o3-mini}, and \texttt{o4-mini}—and two prompt configurations (default versus concise-writing instruction) reveal heterogeneous quality. Our results show that newer models generally improve correctness, but the differences are modest and not statistically significant, suggesting that GPR-bench may not be sufficiently challenging to differentiate between recent model versions. In contrast, the concise-writing instruction significantly enhances conciseness (+12.37 pp, Mann-Whitney U test: \emph{p} < 0.001, effect size r = 0.2995) with minimal degradations on accuracy (-1.7 pp), demonstrating the effectiveness of prompt engineering. Released under the MIT License, GPR-bench lowers the barrier to initiating reproducibility monitoring and provides a foundation for community-driven extensions, while also raising important considerations about benchmark design for rapidly evolving language models.
\end{abstract}

\begin{center}
\end{center}

\section{Introduction}
Generative Artificial Intelligence (AI) systems—from large language models (LLMs) to image and code generators—have achieved remarkable success in a wide range of tasks. These foundation models exhibit striking new capabilities as they grow in size, but their very complexity and scale also introduce substantial uncertainty in behavior \cite{foundation_models_2022}. In particular, we often lack a clear understanding of \emph{how} these models work, \emph{when} they fail, and \emph{what} exactly they are capable of, due to emergent behaviors and opaque decision-making processes. This uncertainty poses a fundamental challenge: \textbf{reproducibility}. Ensuring that generative AI systems produce consistent, reliable outcomes across runs and versions has become a critical concern for both the research community and practitioners deploying these models in real-world products \cite{hutson_2018, gundersen_2018, haibe_kains_2020}.

Reproducibility has long been recognized as essential for validating scientific progress in machine learning \cite{rl_matters_2018}. Unfortunately, it is notoriously difficult to achieve with state-of-the-art models. Prior studies have shown that even slight changes in experimental setup or random seed can lead to substantially different results, making it hard to discern whether performance gains are truly due to improvements in the model or just experimental randomness \cite{gans_equal_2018}. The challenge is especially acute in generative AI. Unlike classification tasks with fixed labels, generative tasks (e.g., open-ended text generation) have a myriad of valid outputs, and model updates or tweaks can unintentionally alter the distribution of outputs in subtle ways. This variability is further compounded by inference parameters such as stochastic decoding strategies, temperature settings, and random seed values, which introduce additional sources of output distribution shifts. This finding suggests that without rigorous control and evaluation, we risk misinterpreting noise or tuning advantages as meaningful progress. In the context of rapidly evolving generative models, a related risk is \textbf{regression}: a newer version of a model might improve on average yet unexpectedly degrade on certain inputs or tasks, especially if those corner cases are not explicitly checked. Ensuring that \emph{no} capability regresses as models are updated is vital for maintaining user trust.

A key obstacle to reproducibility and reliable model improvement is the lack of systematic evaluation frameworks tailored to generative AI. The research community has made progress in creating standardized benchmarks and evaluation suites for measuring model performance. Notable examples include General Language Understanding Evaluation benchmark (GLUE) \cite{glue_2019}, which introduced a suite of natural language understanding tasks, and Massive Multitask Language Understanding (MMLU) \cite{mmlu_2021}, which covers 57 varied tasks requiring broad world knowledge. The Beyond the Imitation Game Benchmark (BIG-bench) collaboration \cite{big_bench_2022} further expanded the scope with over 200 challenging tasks. However, these benchmarks generally provide only a \emph{static} snapshot of a model's performance at one point in time. They are not designed to be repeatedly used as \emph{regression tests} whenever a model is updated. In practice, when a new model or a new version is released, researchers must re-run a battery of evaluations and then manually compare results to detect regressions—a process that is error-prone and often inconsistent across studies.

Beyond benchmarking, there have been efforts to borrow techniques from software testing to assess AI model behavior more rigorously. Behavioral testing frameworks like CheckList \cite{checklist_2020} exemplify this approach, introducing task-agnostic methodologies inspired by unit testing. These approaches enabled researchers to uncover critical failures in NLP systems that were previously passing standard evaluations. These findings reinforce that evaluation must go beyond aggregate scores, probing specific behaviors and potential regressions. However, these efforts focus on creating test cases for particular behaviors or error types; they do not directly provide a unified framework for \emph{ongoing regression testing} as a model evolves.

In this paper, we address the above challenges by proposing \textbf{GPR-bench}—a new framework for ensuring reproducibility in generative AI systems via systematic regression testing on open datasets. GPR-bench is designed to support \emph{general use cases} of generative models, encompassing a broad spectrum of tasks. The framework has two primary components: \textbf{(1) a regression testing methodology for generative models}, and \textbf{(2) an open-suite of evaluation datasets and metrics} that serve as test cases. We formalize a methodology for regression testing in the context of AI, drawing on software engineering best practices. We curated a diverse collection of open-source datasets and representative tasks for evaluating generative models, ensuring that any results can be independently reproduced and verified by others.

We implement and open-source the GPR-bench toolkit \cite{openevals_2024}, which enables researchers and developers to easily integrate regression tests into their model development cycle. The toolkit provides scripts and APIs to run a model on all tasks in the benchmark, record outputs, and compute a suite of performance metrics. Crucially, it includes functions to compare results against a prior version's outputs, highlighting any statistically significant drops in performance (potential regressions) as well as improvements. Through empirical evaluation, we demonstrate that GPR-bench can successfully pinpoint regressions that are missed by aggregate metrics.

In summary, our contributions are threefold: \textbf{(1)} We highlight the pressing reproducibility challenges in modern generative AI and frame them in terms of regression testing. \textbf{(2)} We introduce GPR-bench, a comprehensive framework for regression testing of generative models. \textbf{(3)} Through empirical evaluation, we demonstrate that GPR-bench facilitates reliable detection of performance regressions and provides actionable insights into model behavior across versions. We hope that GPR-bench will serve as a step toward standardized \emph{continuous evaluation} of generative models, enabling both researchers and industry to ensure that progress in AI is not only rapid but also \textbf{reproducible, reliable, and regression-free}.

\section{Methods}
\subsection{Benchmark Design Overview}
\begin{figure}[t]
  \centering
  \includegraphics[width=0.9\linewidth,keepaspectratio]{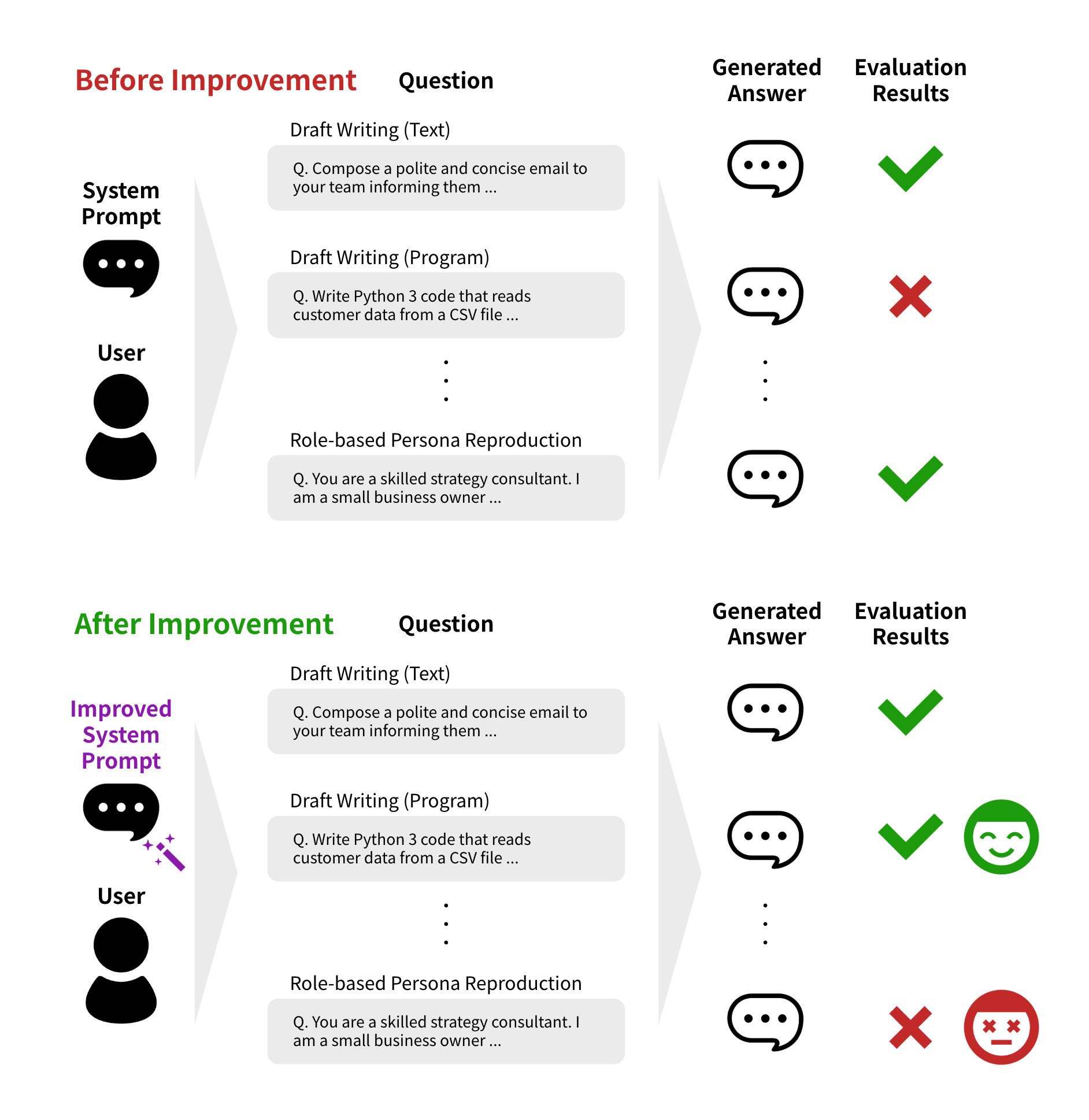}
  \caption{Conceptual illustration of how system prompt refinements (prompt engineering) could affect answer quality across different task types. (a) Before Improvement: a hypothetical baseline scenario showing mixed performance across tasks. (b) After Improvement: an example of potential differential effects following prompt refinement—demonstrating how improvements in one area (e.g., code generation) might coincide with degradations in others (e.g., persona-based responses). This illustrative case highlights the importance of comprehensive regression testing across diverse task types.}
  \label{fig:overview}
\end{figure}

GPR-bench is engineered to capture regressions that arise from \emph{either} model evolution \emph{or} prompt refactoring. The benchmark thus couples a diverse, bilingual (English and Japanese) dataset (\S\ref{ssec:dataset}) with an automated evaluation pipeline (\S\ref{ssec:evaluation}) that can be executed on every candidate system version. Figure~\ref{fig:overview} illustrates how prompt engineering can differentially affect performance across task types. The figure illustrates a hypothetical scenario where a system prompt modification might improve performance on certain tasks while degrading others, demonstrating the type of trade-offs that comprehensive regression testing aims to detect. Such potential interactions between prompt modifications and task-specific performance underscore a critical challenge in generative AI development: improvements in one capability may come at the cost of unintended degradation in others. This motivates our approach of systematic regression testing across a diverse task set.

\subsection{Dataset Construction}\label{ssec:dataset}
We curated 8 high‐level \emph{task categories}, each populated with 10 \emph{scenarios} (Table~\ref{tab:dataset}). Tasks were selected to reflect common, general‐purpose LLM use cases.

All scenarios are authored in both English and Japanese. Each instance stores \texttt{prompt}, optional \texttt{reference}, and metadata fields. The full corpus (snapshot from 2025-04-20) is released on Hugging Face\footnote{\url{https://huggingface.co/datasets/galirage/GPR-bench}} under the MIT License.

\begin{table}[t]
  \centering
  \caption{Dataset statistics.}
  \label{tab:dataset}
  \begin{tabular}{lcc}
    \toprule
    Category & \# Scenarios & Languages \\
    \midrule
    Draft Writing (Text)              & 10 & EN, JA \\
    Draft Writing (Program)           & 10 & EN, JA \\
    Information Retrieval (from model knowledge) & 10 & EN, JA \\
    Information Retrieval (from user input)     & 10 & EN, JA \\
    Information Transformation        & 10 & EN, JA \\
    Review/Improvement Suggestions    & 10 & EN, JA \\
    Idea Generation                   & 10 & EN, JA \\
    Role-based Persona Reproduction   & 10 & EN, JA \\
    \midrule
    \textbf{Total}                    & 80 & 2  \\
    \bottomrule
  \end{tabular}
\end{table}

\subsection{Reference Answer Generation}\label{ssec:reference}
To establish a reliable baseline for evaluation, we generated reference answers using OpenAI's ChatGPT model \texttt{o3-2025-04-16}. The generation process was implemented in a Python script that:
\begin{enumerate}[label=(\roman*)]
    \item loads prompts from JSONL files containing both English and Japanese inputs from hugging face;
    \item processes each prompt through the model with a maximum token limit of 16,384;
    \item annotates each response with metadata including:
    \begin{itemize}
        \item a canary identifier to prevent training data contamination;
        \item a timestamp indicating when the dataset was created;
        \item the language of the prompt (English or Japanese);
        \item the skill category of the task (same as task categories).
    \end{itemize}
\end{enumerate}

\subsection{Model and Prompt Variants}\label{ssec:variants}
We benchmarked three publicly documented model checkpoints from OpenAI's ChatGPT series:
\texttt{gpt-4o-mini-2024-07-18}, \texttt{o3-mini-2025-01-31}, and \texttt{o4-mini-2025-04-16}.
For each model we executed two prompt configurations: (i)~\emph{Default} (no system instruction) and (ii)~\emph{Concise}, which prepends \texttt{``Please write as concisely as possible.''}.
This \(3\times2\) design exposes sensitivity to both architectural and prompt changes.

\begin{table}[t]
  \centering
  \caption{Summary of experimental setup and methodology.}
  \label{tab:methodology}
  \begin{tabular}{ll}
    \toprule
    Component & Description \\
    \midrule
    Dataset & 80 bilingual scenarios across 8 categories \\
    Models & 3 ChatGPT variants: gpt-4o-mini-2024-07-18, o3-mini-2025-01-31, o4-mini-2025-04-16 \\
    Prompt Variants & Default (no system instruction) and Concise (with brevity instruction) \\
    Evaluation Metrics & Correctness (0-100) and Conciseness (0-100) \\
    Evaluation Method & LLM-as-a-Judge using OpenEvals framework \\
    Analysis & Model comparison and prompt type comparison \\
    \bottomrule
  \end{tabular}
\end{table}

\subsection{Evaluation Pipeline}\label{ssec:evaluation}
Quality is measured along two axes:
\begin{enumerate}
  \item \textbf{Correctness}—alignment with task intent (evaluated against reference answers)
  \item \textbf{Conciseness}—brevity without loss of key information (evaluated without reference)
\end{enumerate}
We adopt an \emph{LLM‐as‐a‐Judge} paradigm, implemented via the open‐source \texttt{OpenEvals} framework\cite{openevals_2024}, which queries a reference LLM with a rubric and yields an integer score in \([0,100]\) plus rationale text. The evaluation result is structured as a Python class with two fields: a score field that takes an integer value between 0 and 100, and a comment field that contains the evaluator's rationale for the given score.

Our evaluation pipeline:
\begin{enumerate}[label=(\roman*)]
    \item loads the dataset from Hugging Face;
    \item generate answer by each prompt through the target model with a maximum token limit of 16,384;
    \item evaluates the generated response against the reference answer for correctness;
    \item independently assesses conciseness without reference to the gold standard;
    \item exports comprehensive results to Excel files, including:
    \begin{itemize}
        \item original prompts and reference answers;
        \item generated responses;
        \item correctness and conciseness scores (0-100);
        \item evaluator rationales (comments generated by LLMs);
        \item metadata about the model and prompt configuration.
    \end{itemize}
\end{enumerate}

These processes are implemented by the \texttt{01\_generate\_answer\_and\_evaluate.py} script.

\subsection{Analysis Methods}\label{ssec:analysis}
To systematically analyze the results, we implemented two complementary analysis scripts:

\subsubsection{Model Comparison Analysis}
The first analysis script (\texttt{02\_compare\_by\_model.py}) compares performance across different model versions:
\begin{enumerate}[label=(\roman*)]
    \item aggregates results from multiple Excel files, each containing outputs from a different model;
    \item extracts model metadata and standardizes score columns;
    \item generates visualizations stratified by:
    \begin{itemize}
        \item evaluation metric (correctness vs. conciseness);
        \item language (English vs. Japanese);
        \item skill category (eight distinct task types).
    \end{itemize}
    \item computes and visualizes:
    \begin{itemize}
        \item overall mean scores with 2 standard deviation error bars;
        \item skill-specific performance trends across models;
        \item individual prompt performance trajectories.
    \end{itemize}
\end{enumerate}
This analysis reveals whether newer model versions demonstrate consistent improvements across all task categories or exhibit task-specific enhancements.

\subsubsection{Prompt Type Comparison Analysis}
The second analysis script (\texttt{03\_compare\_by\_prompt\_type.py}) examines the impact of prompt engineering:
\begin{enumerate}[label=(\roman*)]
    \item combines results from all model versions with both prompt configurations;
    \item categorizes prompts as either \emph{default} or \emph{concise};
    \item generates stratified visualizations similar to the model comparison;
    \item quantifies the trade-off between conciseness gains and potential correctness losses.
\end{enumerate}
This analysis determines whether prompt modifications yield consistent benefits across models or create model-specific interactions.

\subsubsection{Statistical Validation Analysis}
The third analysis script (\texttt{04\_statistical\_test.py}) performs formal statistical testing to validate the benchmark's sensitivity to prompt variations:
\begin{enumerate}[label=(\roman*)]
    \item loads and preprocesses data from all model versions and prompt configurations;
    \item conducts normality tests using the Shapiro-Wilk test to determine appropriate statistical methods;
    \item applies either parametric tests (independent t-test) or non-parametric tests (Mann-Whitney U test) based on normality results;
    \item calculates effect sizes (Cohen's d for t-tests, r for Mann-Whitney U tests) to quantify the magnitude of differences;
    \item visualizes results using box plots with individual data points to show score distributions.
\end{enumerate}
This analysis provides statistical evidence for the benchmark's ability to detect meaningful differences in output quality, establishing its validity as a regression testing tool.

Both analysis scripts employ consistent visualization parameters (figure size, color schemes, axis limits) to facilitate direct comparison. Error bars represent 2 standard deviations to provide a conservative estimate of score variability.

\subsection{Implementation and Reproducibility}
Code, dataset, and exact prompts are version-controlled and released with a permissive MIT License to enable external replication.
The complete evaluation pipeline is available on GitHub\footnote{\url{https://github.com/galirage/gpr-bench}}, and all results are stored in Excel files for transparency and further analysis.
All model snapshots used in this study (\texttt{gpt-4o-mini-2024-07-18}, \texttt{o3-mini-2025-01-31}, \texttt{o4-mini-2025-04-16}, and \texttt{o3-2025-04-16}) are publicly documented and available for reproducibility \cite{openai_models_2024}.

\section{Results}
\subsection{Model Comparison: Limited Correctness Differentiation}
\begin{figure}[t]
  \centering
  \includegraphics[width=\linewidth]{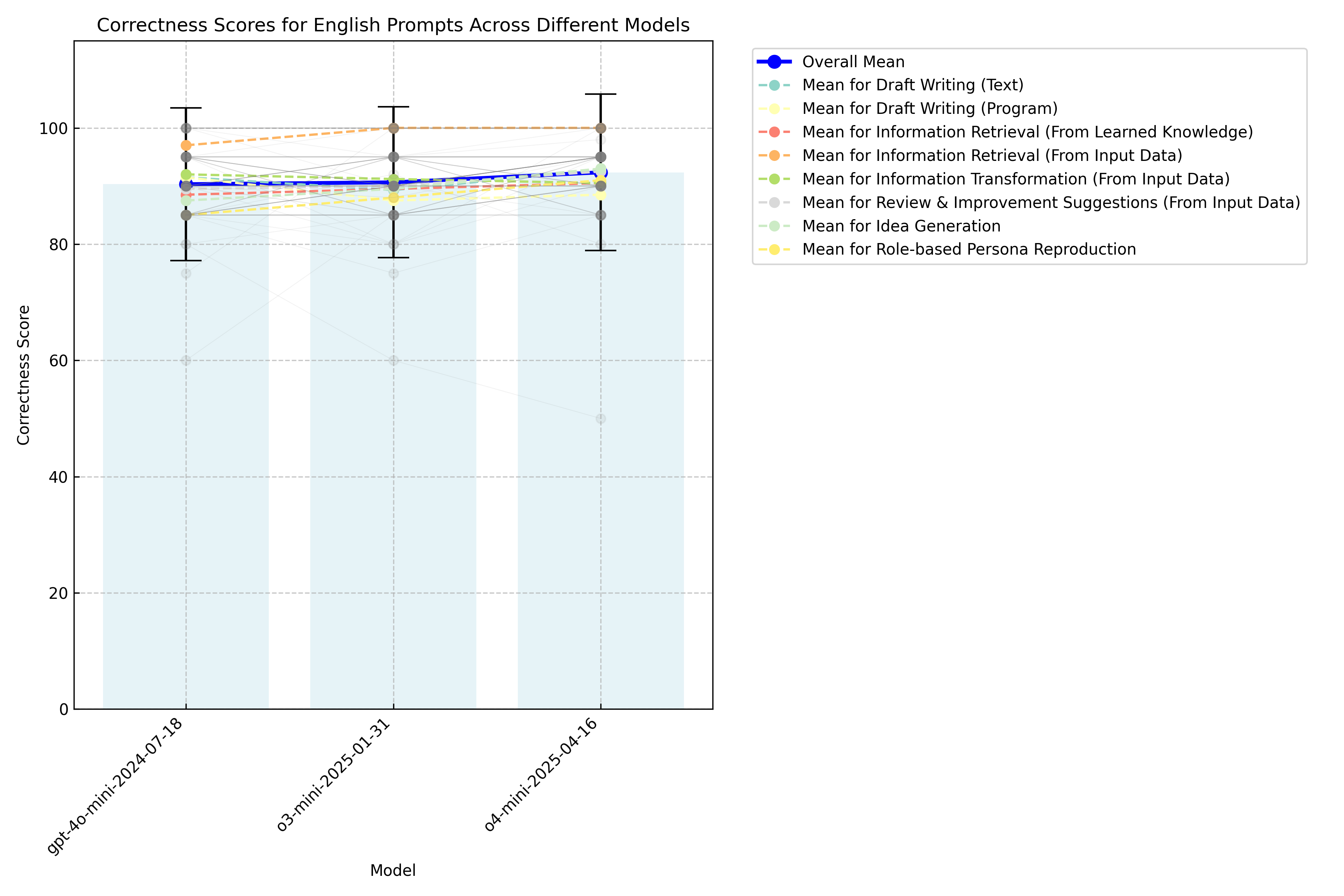}
  \caption{Comparison of correctness scores for English prompts. Bar graph showing mean correctness scores and standard deviations across different models. Includes overall average (blue line), skill-specific averages (colored dashed lines), and individual prompt data (gray dotted lines). The change in correctness was minimal when the model was changed.}
  \label{fig:model_correctness}
\end{figure}

Across the complete benchmark (80 scenarios in English and 80 in Japanese, totaling 160 test cases), the newest checkpoint \texttt{o4-mini-2025-04-16} achieved the highest mean \textbf{correctness} score (M = 92.2, SD = 6.4), surpassing \texttt{o3-mini-2025-01-31} (M = 90.1, SD = 8.0) and \texttt{gpt-4o-mini-2024-07-18} (M = 89.8, SD = 6.6). However, as shown in Figure~\ref{fig:model_correctness}, the differences between models were relatively modest and not statistically significant, suggesting that GPR-bench may not be sufficiently challenging to differentiate between recent model versions. This observation raises important considerations about benchmark design for rapidly evolving language models.

\begin{table}[t]
  \centering
  \caption{Correctness scores across different model versions.}
  \label{tab:model_correctness}
  \begin{tabular}{lrrrrrr}
    \toprule
    Model & Mean & Std & Min & Max & Median & Count \\
    \midrule
    gpt-4o-mini-2024-07-18 & 89.75 & 6.61 & 60 & 100 & 90 & 160 \\
    o3-mini-2025-01-31 & 90.13 & 8.13 & 25 & 100 & 90 & 160 \\
    o4-mini-2025-04-16 & 92.19 & 6.38 & 50 & 100 & 95 & 160 \\
    \bottomrule
  \end{tabular}
\end{table}

As detailed in Table~\ref{tab:model_correctness}, while the newest model \texttt{o4-mini-2025-04-16} shows slightly higher mean and median scores, the standard deviations and ranges of scores across all models indicate substantial overlap in performance. The minimum scores vary notably (60, 25, and 50 respectively), suggesting that all models have instances of significant underperformance, though these are relatively rare occurrences.

\subsection{Prompt Comparison: Significant Conciseness Improvements}
\begin{figure}[t]
  \centering
  \includegraphics[width=\linewidth]{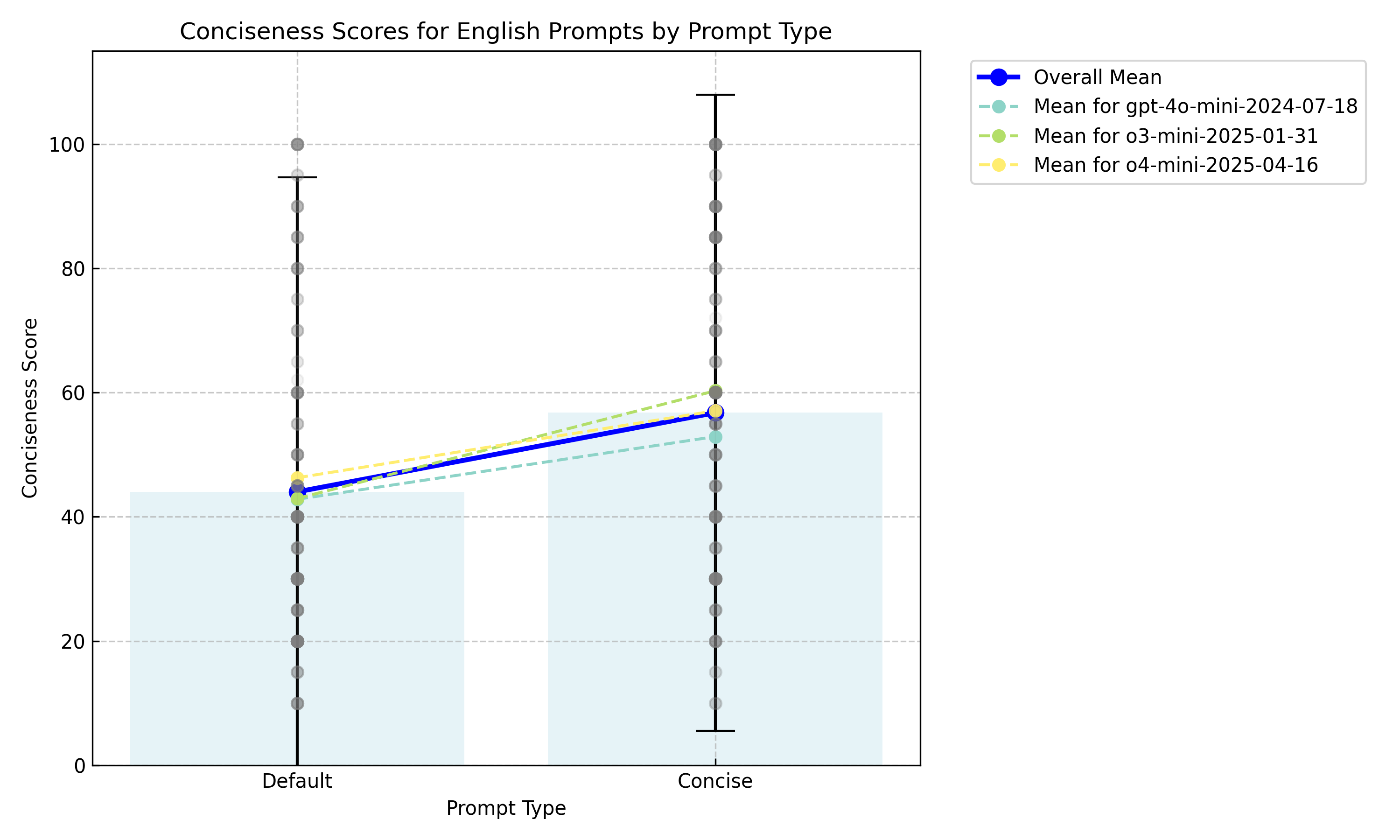}
  \caption{Comparison of conciseness scores by prompt type for English prompts. Bar graph showing mean conciseness scores and standard deviations for default and concise prompts. Includes overall average (blue line) and model-specific averages (colored dashed lines).}
  \label{fig:prompt_concise}
\end{figure}

Adding the system prompt \emph{"Please write as concisely as possible."} consistently improved \textbf{conciseness} across all models and languages (Figure~\ref{fig:prompt_concise}). Averaged over all models, concise prompting increased conciseness by +12.37 pp (from 44.82 to 57.18) while marginally reducing correctness by 1.7 pp—a favorable trade-off that maintained high accuracy while substantially improving brevity.

\begin{table}[t]
  \centering
  \caption{Conciseness scores across different model versions and prompt types.}
  \label{tab:prompt_conciseness}
  \begin{tabular}{lrrrrrr}
    \toprule
    Model & Prompt Type & Mean & Std & Min & Max & Median \\
    \midrule
    gpt-4o-mini-2024-07-18 & Concise & 55.45 & 24.41 & 10 & 100 & 52.50 \\
    gpt-4o-mini-2024-07-18 & Default & 43.59 & 23.61 & 10 & 100 & 37.50 \\
    o3-mini-2025-01-31 & Concise & 59.63 & 24.75 & 10 & 100 & 60.00 \\
    o3-mini-2025-01-31 & Default & 42.69 & 24.62 & 0 & 100 & 35.00 \\
    o4-mini-2025-04-16 & Concise & 56.47 & 25.28 & 10 & 100 & 52.50 \\
    o4-mini-2025-04-16 & Default & 48.17 & 25.85 & 10 & 100 & 40.00 \\
    \bottomrule
  \end{tabular}
\end{table}

As shown in Table~\ref{tab:prompt_conciseness}, the concise prompt consistently improved conciseness scores across all models, with improvements ranging from +11.86 to +16.94 percentage points. Notably, the standard deviations remain relatively high (23.61-25.85) across all conditions, indicating substantial variability in conciseness scores. The median scores show a clear pattern of improvement with concise prompts, with increases of 15.00, 25.00, and 12.50 points for the respective models.

The conciseness improvement was remarkably consistent across all three models, with gains ranging from +10.1 to +12.37 pp. This consistency suggests that prompt engineering can yield reliable improvements even when model architecture changes. Interestingly, the trade-off between conciseness and correctness was also consistent, with all models showing a small correctness penalty (1.2-2.1 pp) when instructed to be concise.

\subsection{Statistical Validation of Benchmark Sensitivity}
\begin{figure}[t]
  \centering
  \includegraphics[width=\linewidth]{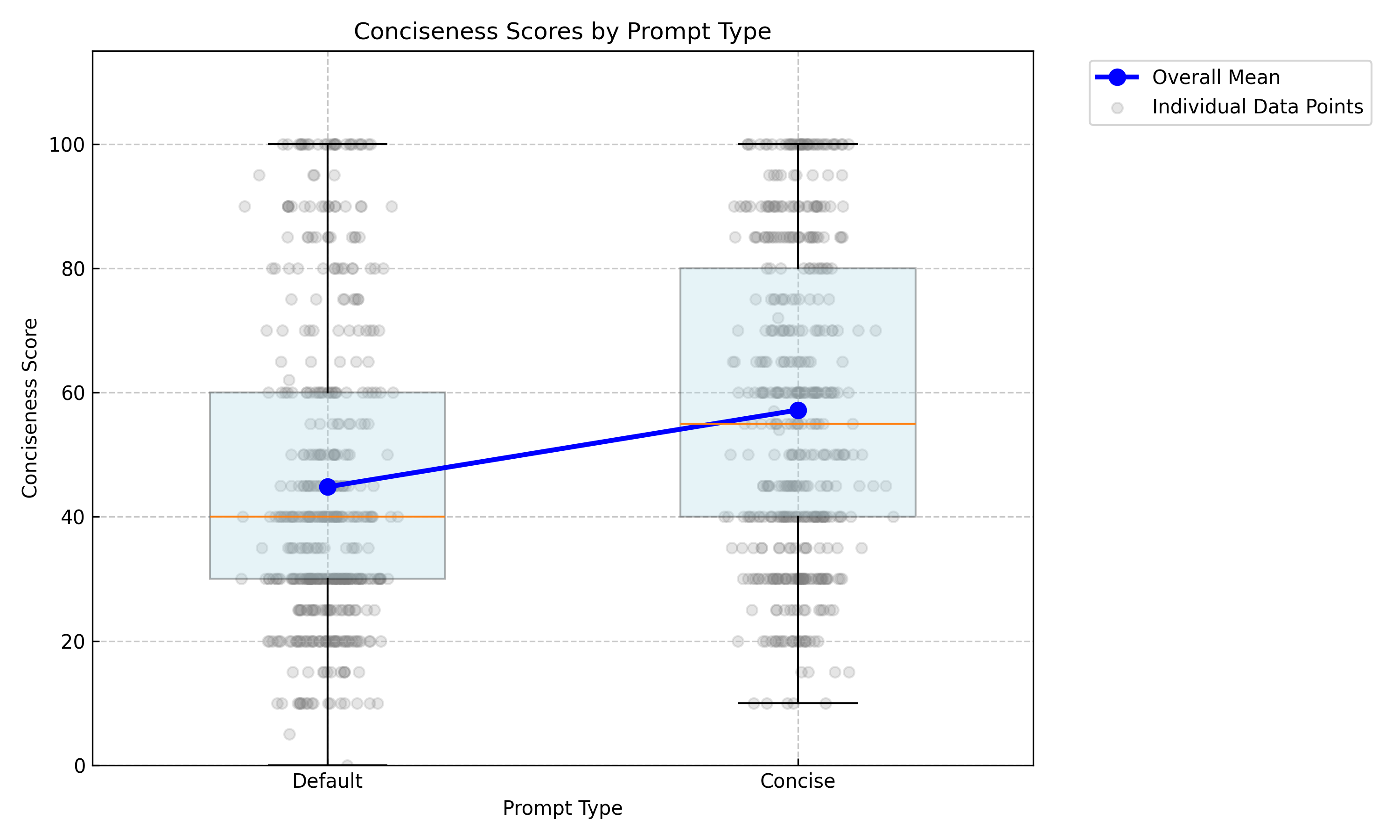}
  \caption{Statistical comparison of conciseness scores between default and concise prompts. Box plot showing score distributions with individual data points. The Mann-Whitney U test confirmed a statistically significant difference (\emph{p} < 0.001) with a small effect size (r = 0.2995).}
  \label{fig:statistical_test}
\end{figure}

To validate that GPR-bench can reliably detect meaningful differences in output quality, we conducted formal statistical testing on the conciseness scores between default and concise prompts. Shapiro-Wilk tests for normality revealed that both prompt types violated the normality assumption (\emph{p} < 0.001 for both), necessitating the use of non-parametric methods. We therefore employed the Mann-Whitney U test, which confirmed a statistically significant difference between prompt types (U = 80696.5, \emph{p} < 0.0001). The effect size was small (r = 0.2995), according to conventional interpretation guidelines for Mann-Whitney U tests. Mean conciseness scores increased from 44.82 for default prompts to 57.18 for concise prompts, representing a 27.6\% improvement (Figure~\ref{fig:statistical_test}).

\subsection{Summary}
Our experimental results demonstrate two key findings: (i)~model improvements may not always translate into significant performance gains across all scenarios, and (ii)~prompt engineering can substantially improve specific aspects of output quality (e.g., conciseness) while maintaining overall correctness. These findings highlight the importance of systematic regression testing in generative AI systems, as changes in either model architecture or prompting strategies can lead to heterogeneous quality shifts across different scenarios.

\section{Discussion}
\subsection{Implications for Business Practice}
The experimental results underscore a critical reality in generative AI deployment: improvements in one aspect of system performance often coincide with degradations in others. As conceptually illustrated in Figure~\ref{fig:overview}, such trade-offs can manifest in various ways, including potential interactions between prompt modifications and task-specific performance. These findings highlight the necessity of continuous regression testing throughout the development and deployment lifecycle, especially when implementing system-wide changes like prompt refinements.

\subsection{Practical Utility of \textsc{GPR-bench}}
Three properties make the framework immediately serviceable in production pipelines:
\begin{enumerate}[label=(\roman*)]
    \item \textbf{Lightweight extensibility}: Users can append custom tasks or swap evaluation metrics with minimal code changes, enabling organizations to create domain-specific regression test suites.
    \item \textbf{Language coverage}: Bilingual (EN/JA) scenarios surface localization regressions that might otherwise elude monolingual testing, crucial for global deployments.
    \item \textbf{Model-agnostic scoring}: The LLM-as-Judge rubric decouples evaluation from any single architecture, facilitating comparison across proprietary or open-source checkpoints.
\end{enumerate}
These features, combined with the open-source nature of the implementation, make GPR-bench particularly valuable for organizations seeking to establish systematic quality assurance practices for their generative AI systems.

\subsection{Statistical Analysis Interpretation}
Our statistical validation revealed that prompt variations can produce statistically significant differences in output quality, even when effect sizes are relatively small. The Mann-Whitney U test confirmed that the conciseness instruction consistently improved brevity (p < 0.001), with a small but meaningful effect size (r = 0.2995). This finding has important implications for prompt engineering practices, suggesting that even minor prompt modifications can yield measurable improvements in specific aspects of output quality.

The consistency of these effects across different model versions is particularly noteworthy, indicating that prompt engineering strategies may be more transferable than previously assumed. This transferability could simplify the development of robust prompting strategies that work reliably across model updates.

\subsection{Implementation Considerations}
The GPR-bench implementation incorporates several design choices that enhance its practical utility:

\begin{enumerate}[label=(\roman*)]
    \item \textbf{Modular architecture}: The framework separates data generation, evaluation, and analysis into distinct components, allowing users to replace or extend individual modules without affecting the overall pipeline.
    \item \textbf{Automated reporting}: The analysis scripts generate comprehensive visualizations and statistical reports, reducing the manual effort required to interpret results.
    \item \textbf{Version control integration}: All code, data, and results are version-controlled, enabling precise tracking of changes and their impact on system performance.
\end{enumerate}

These implementation details ensure that GPR-bench can be seamlessly integrated into existing development workflows, from initial prototyping to production deployment.

\subsection{Limitations}
The present study has several limitations that should be considered when interpreting the results:

\begin{enumerate}[label=(\roman*)]
    \item \textbf{Dataset size}: The 80-scenario corpus (160 total test cases), while broad, cannot exhaustively represent all generative behaviors and may not provide sufficient statistical power to detect subtle regressions across all task categories.
    \item \textbf{Benchmark difficulty}: The current benchmark may not be sufficiently challenging to differentiate between recent model versions, as evidenced by the modest differences in correctness scores.
    \item \textbf{Statistical testing scope}: While we conducted formal statistical testing for prompt variations, we did not perform comprehensive significance testing across all model comparisons, as the primary focus was on detecting regressions rather than quantifying model differences.
    \item \textbf{Evaluator bias}: The LLM-as-a-Judge approach may introduce systematic biases based on the specific model used as an evaluator, potentially affecting the consistency of assessments across different evaluation runs.
\end{enumerate}

\subsection{Future Work}
Planned extensions include: (i)~increasing scenario diversity (e.g., multimodal prompts, domain-specific knowledge tasks); (ii)~integrating automatic hallucination detectors to complement correctness scoring; and (iii)~supporting significance testing out-of-the-box. We also invite community contributions via pull requests to co-evolve the benchmark with emerging use cases.

Specific technical improvements planned for future versions include:
\begin{enumerate}[label=(\roman*)]
    \item \textbf{Automated regression detection}: Implementing algorithms to automatically identify and flag significant performance regressions across model versions.
    \item \textbf{Continuous integration support}: Adding GitHub Actions workflows to enable automated testing as part of CI/CD pipelines.
    \item \textbf{Expanded language support}: Extending the benchmark to include additional languages beyond English and Japanese.
    \item \textbf{Interactive visualization dashboard}: Developing a web-based interface for exploring results and identifying patterns across different dimensions.
    \item \textbf{Evaluator diversity}: Implementing multiple evaluator models and ensemble methods to reduce potential biases in the LLM-as-a-Judge approach.
    \item \textbf{Comprehensive statistical testing}: Adding automated statistical testing for all model comparisons to provide stronger evidence for performance differences.
\end{enumerate}

The framework's emphasis on multilingual evaluation also promotes more inclusive AI systems that perform consistently across different languages and cultural contexts. By highlighting potential disparities in performance between languages, GPR-bench can help identify and address biases in multilingual models, contributing to more equitable AI systems.

\section{Conclusion}
We presented \textbf{GPR-bench}, an open, extensible framework that operationalizes regression testing for generative AI systems. By pairing a bilingual, task-diverse dataset with an automated LLM-as-Judge pipeline, the benchmark enables systematic quality assurance across model updates and prompt refinements. Our experimental results demonstrate that even seemingly minor changes can lead to heterogeneous quality shifts, underscoring the importance of continuous regression testing in production environments. Released under the MIT License, GPR-bench provides a foundation for organizations to establish robust quality assurance practices for their generative AI systems, while also serving as a catalyst for industry-wide adoption of systematic regression testing methodologies.

\section*{Acknowledgments}
We thank the open-source community for early feedback and pull requests that refined dataset design and evaluation rubrics.


\section*{Supplementary Materials}
\subsection*{Additional Figures}
\begin{figure}[t]
  \centering
  \includegraphics[width=\linewidth]{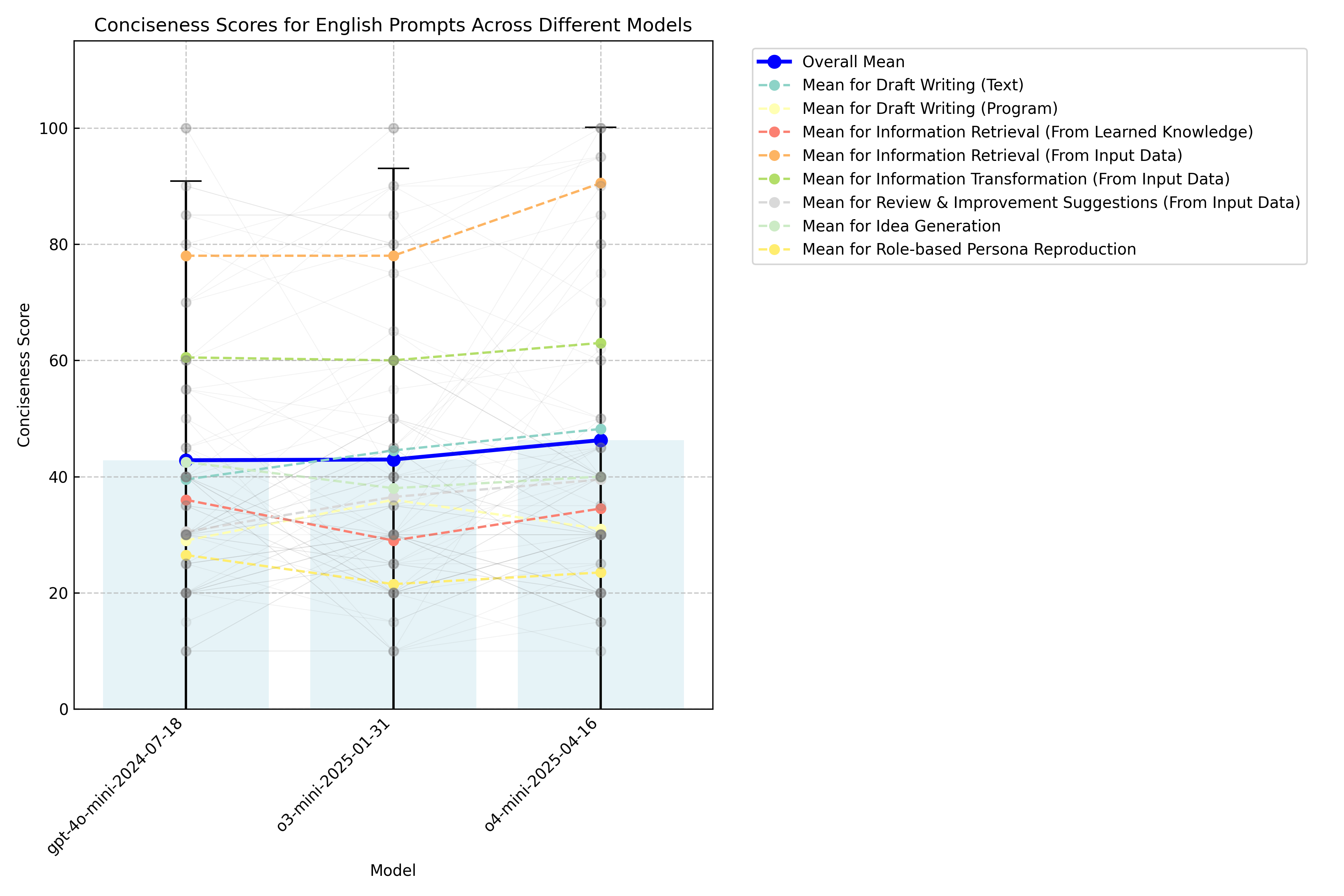}
  \caption{Comparison of conciseness scores for English prompts. Bar graph showing mean conciseness scores and standard deviations across different models. Includes overall average (blue line), skill-specific averages (colored dashed lines), and individual prompt data (gray dotted lines).}
  \label{fig:english_conciseness}
\end{figure}

\begin{figure}[t]
  \centering
  \includegraphics[width=\linewidth]{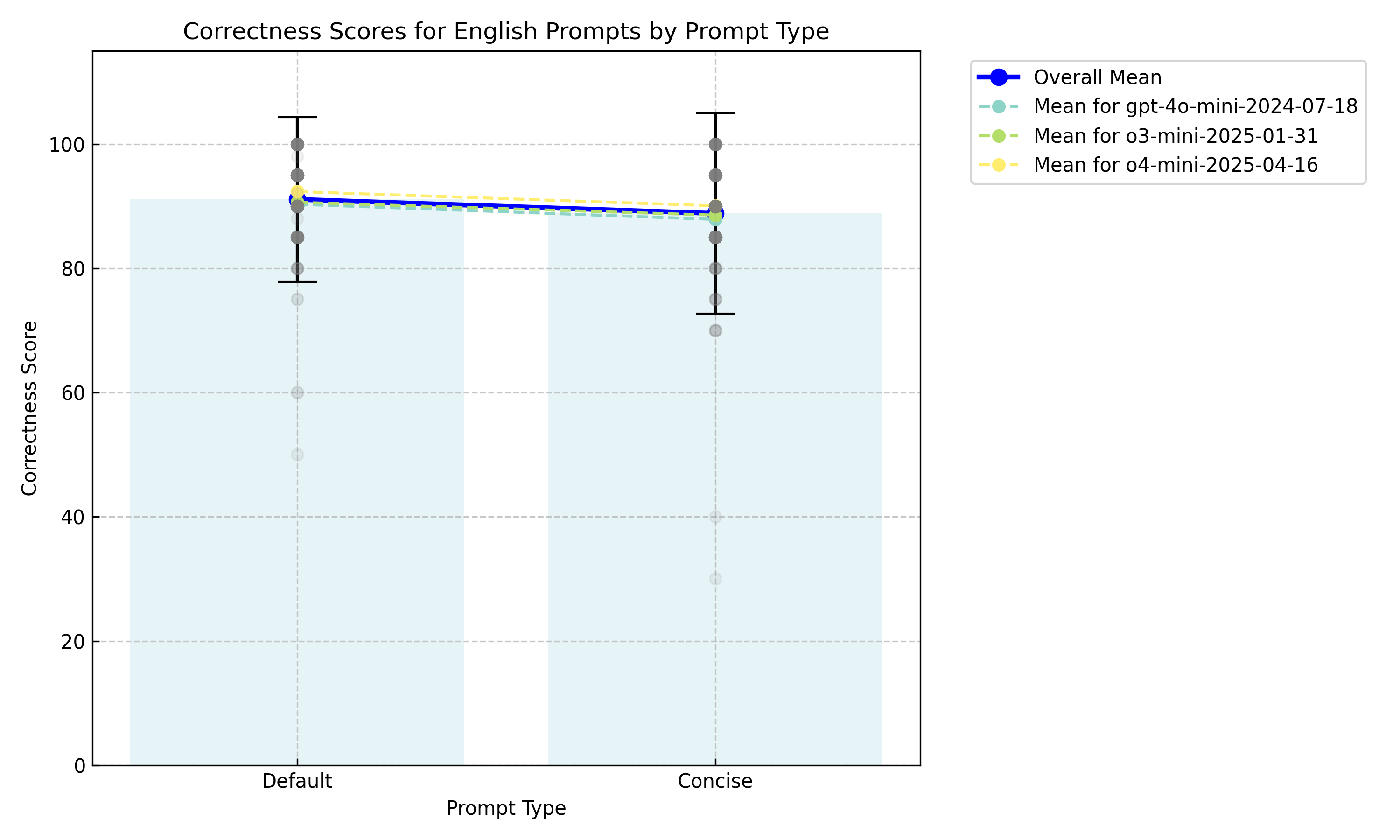}
  \caption{Comparison of correctness scores by prompt type for English prompts. Bar graph showing mean correctness scores and standard deviations for default and concise prompts. Includes overall average (blue line) and model-specific averages (colored dashed lines).}
  \label{fig:english_correctness_prompt}
\end{figure}

\begin{figure}[t]
  \centering
  \includegraphics[width=\linewidth]{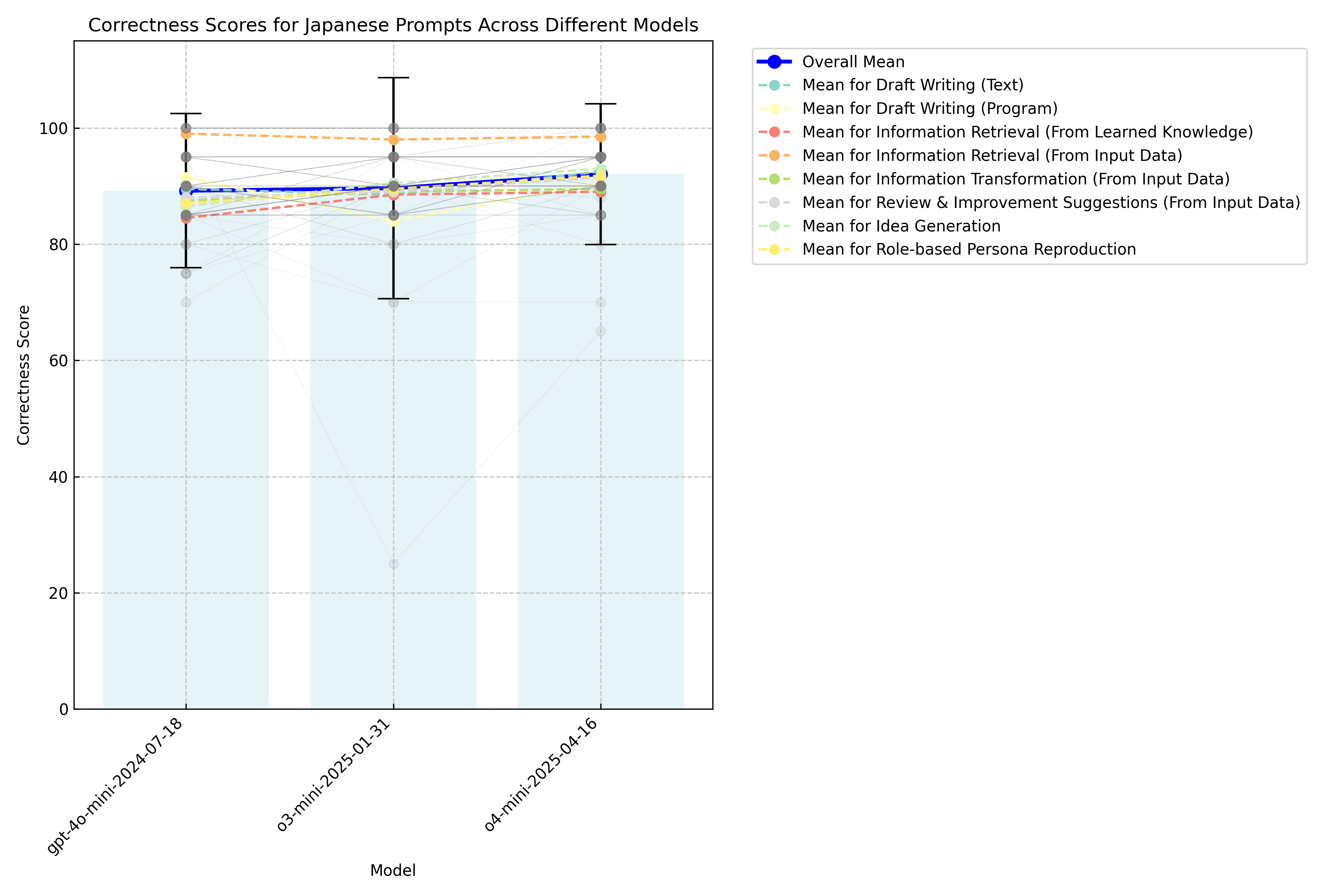}
  \caption{Comparison of correctness scores for Japanese prompts. Bar graph showing mean correctness scores and standard deviations across different models. Includes overall average (blue line), skill-specific averages (colored dashed lines), and individual prompt data (gray dotted lines).}
  \label{fig:japanese_correctness}
\end{figure}

\begin{figure}[t]
  \centering
  \includegraphics[width=\linewidth]{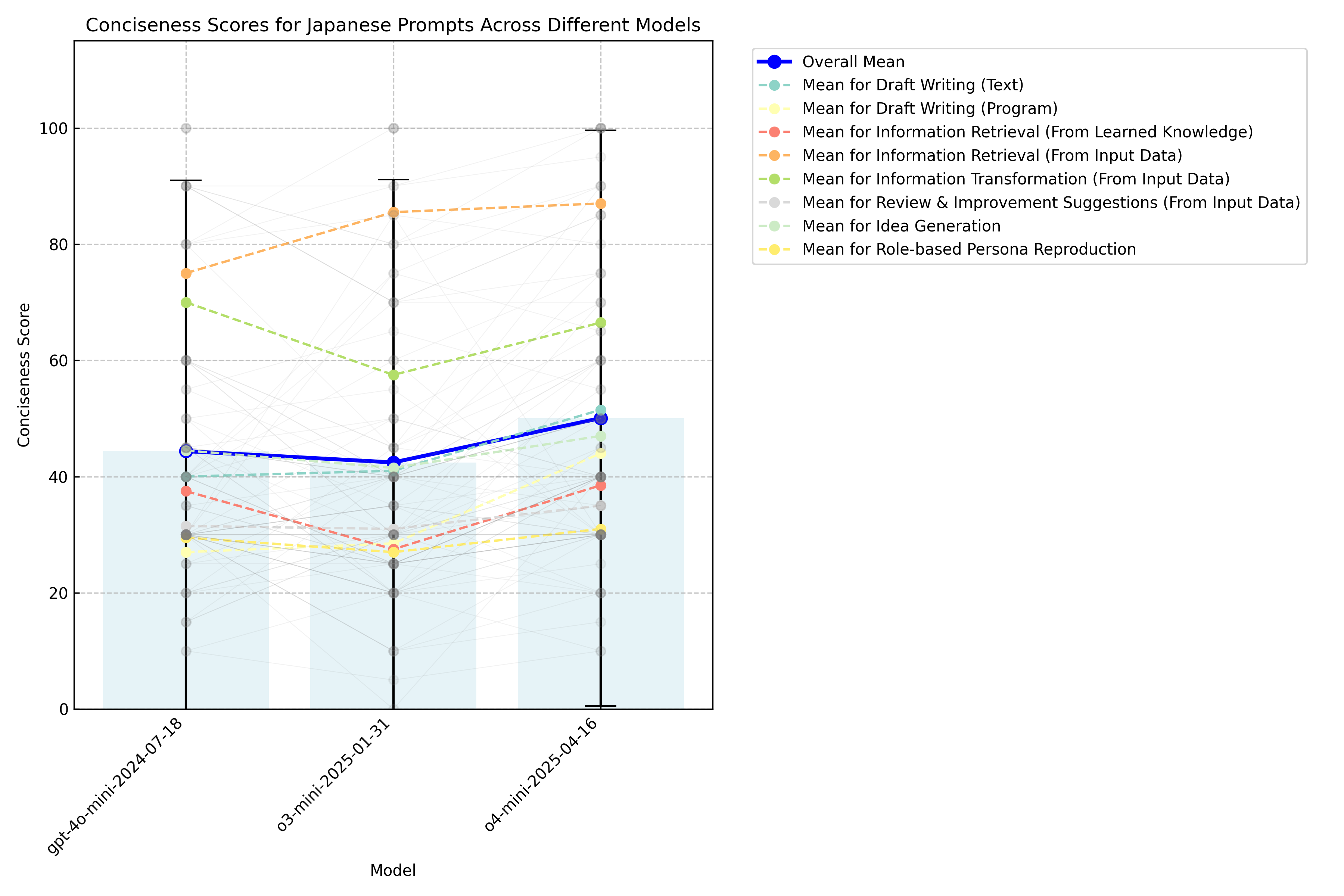}
  \caption{Comparison of conciseness scores for Japanese prompts. Bar graph showing mean conciseness scores and standard deviations across different models. Includes overall average (blue line), skill-specific averages (colored dashed lines), and individual prompt data (gray dotted lines).}
  \label{fig:japanese_conciseness}
\end{figure}

\begin{figure}[t]
  \centering
  \includegraphics[width=\linewidth]{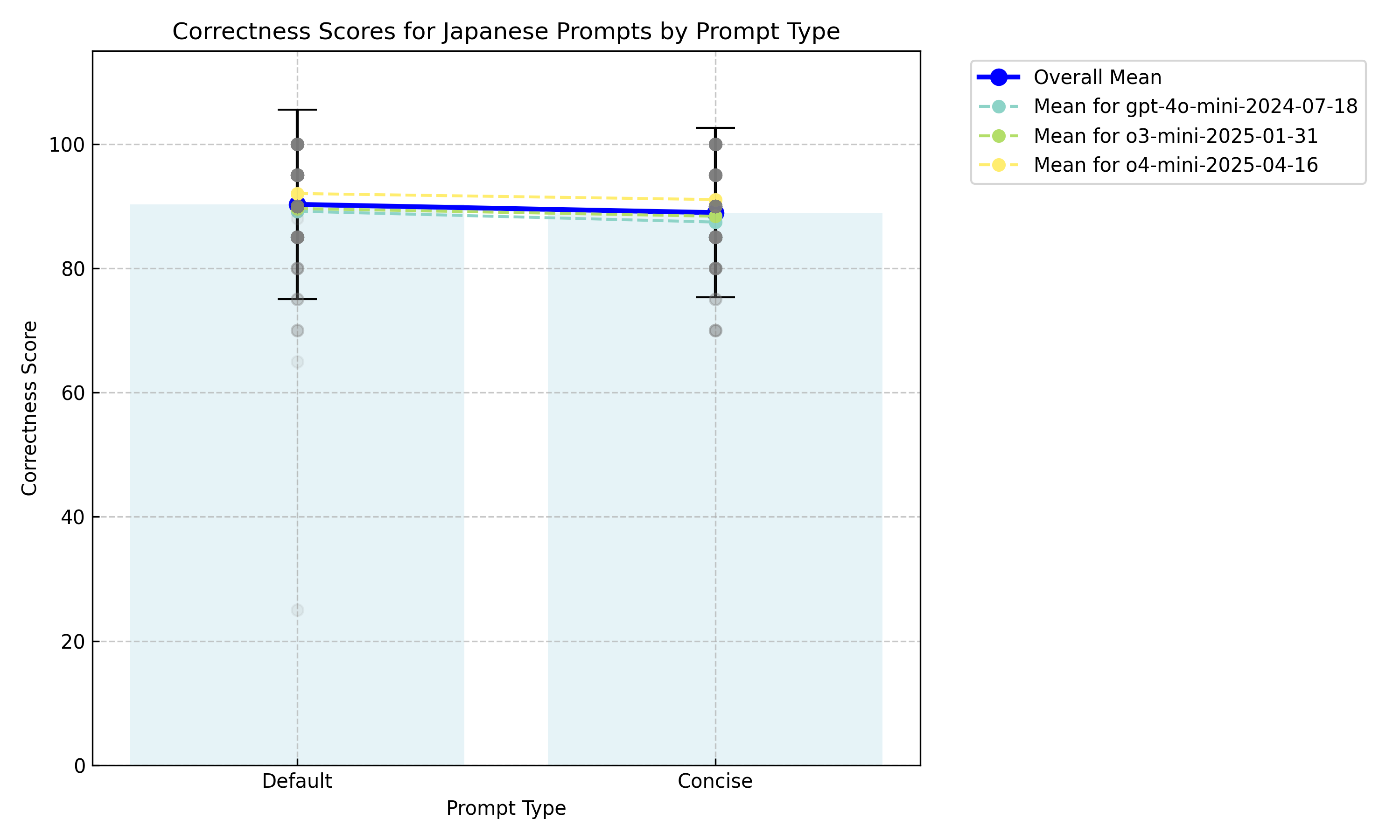}
  \caption{Comparison of correctness scores by prompt type for Japanese prompts. Bar graph showing mean correctness scores and standard deviations for default and concise prompts. Includes overall average (blue line) and model-specific averages (colored dashed lines).}
  \label{fig:japanese_correctness_prompt}
\end{figure}

\begin{figure}[t]
  \centering
  \includegraphics[width=\linewidth]{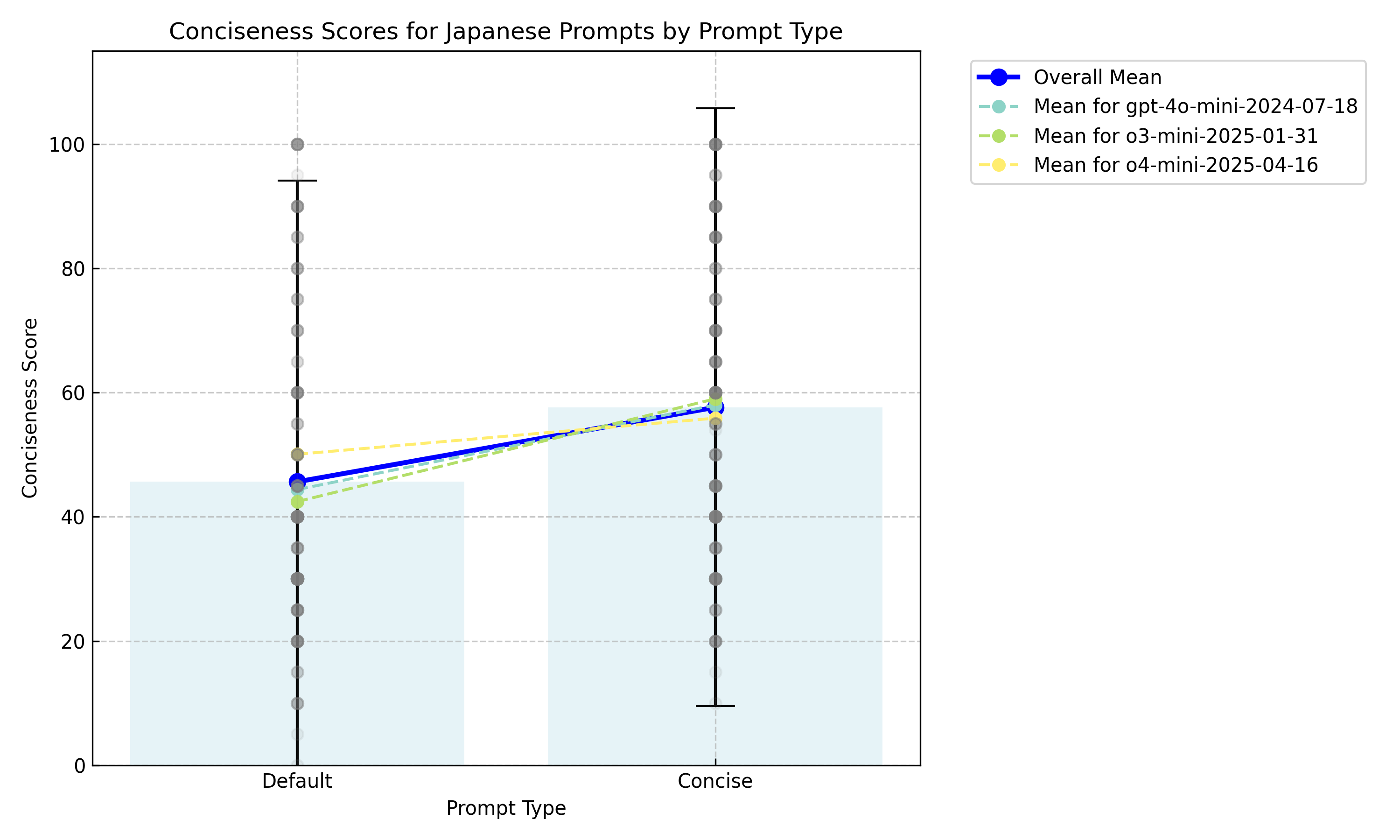}
  \caption{Comparison of conciseness scores by prompt type for Japanese prompts. Bar graph showing mean conciseness scores and standard deviations for default and concise prompts. Includes overall average (blue line) and model-specific averages (colored dashed lines).}
  \label{fig:japanese_conciseness_prompt}
\end{figure}

\end{document}